\documentclass[runningheads]{llncs}
\usepackage[T1]{fontenc}
\usepackage{amsmath}
\usepackage{graphicx}
\usepackage{booktabs}
\usepackage{amssymb}
\usepackage[maxbibnames=99]{biblatex}
\DeclareMathOperator{\FSDEM}{FSDEM}
\DeclareMathOperator{\STAB}{S}
\DeclareMathOperator{\CLACC}{CLACC}
\DeclareMathOperator{\BFI}{BFI}
\begin{document}
\title{FSDEM: Feature Selection Dynamic Evaluation Metric}
%
%\titlerunning{Abbreviated paper title}
% If the paper title is too long for the running head, you can set
% an abbreviated paper title here
%
\author{Muhammad Rajabinasab, Anton D. Lautrup, Tobias Hyrup, Arthur Zimek}
\authorrunning{M. Rajabinasab, A. D. Lautrup, T. Hyrup, A. Zimek}
% First names are abbreviated in the running head.
% If there are more than two authors, 'et al.' is used.
%
\institute{University of Southern Denmark, Odense, Denmark}

\maketitle              % typeset the header of the contribution
\begin{abstract}
Expressive evaluation metrics are indispensable for informative experiments in all areas, and while several metrics are established in some areas, in others, such as feature selection, only indirect or otherwise limited evaluation metrics are found. In this paper, we propose a novel evaluation metric to address several problems of its predecessors and allow for flexible and reliable evaluation of feature selection algorithms. The proposed metric is a dynamic metric with two properties that can be used to evaluate both the performance and the stability of a feature selection algorithm. We conduct several empirical experiments to illustrate the use of the proposed metric in the successful evaluation of feature selection algorithms. We also provide a comparison and analysis to show the different aspects involved in the evaluation of the feature selection algorithms. The results indicate that the proposed metric is successful in carrying out the evaluation task for feature selection algorithms.
\\~\\
\textit{This paper is an extended version of a paper published at SISAP 2024.}
\end{abstract}
\section{Introduction}
Many novel algorithms are proposed each year to solve different difficult and complex problems. In some cases, newly proposed algorithms outperform previous methods in every aspect and in all instances of the problem, while more often, this is not the case \cite{DBLP:journals/algorithms/MostertME21}. Novel algorithms either show superior figures only in some instances and aspects of the problem, or they are not generalizable to other applications or domains. This can be due to insufficient experiments, incomplete analysis, or the usage of evaluation metrics which are not able to accurately reflect different performance aspects of the algorithm.

Feature selection is the task of selecting the most informative and important features for the target machine learning or data mining task and removing redundant features. Feature selection is usually employed in a supervised setting, where either statistical values are calculated as feature importance (filter methods) or feature importance is assessed based on the impact of the feature on a classifier's prediction (wrapper methods) \cite{DBLP:journals/apin/DhalA22}, whereas recently unsupervised feature selection methods have gained attention as they are able to select the most informative features without requiring the use of labels \cite{10123301, DBLP:journals/ijon/ShangKWZWLJ23, DBLP:journals/prl/WANG2024110183}.
Feature selection takes a global approach to select a subset of representational features for a full dataset as opposed to local approaches such as subspace methods for clustering \cite{HouKieZim23} or outlier detection \cite{DBLP:journals/sadm/ZimekSK12}, or to projection and approximation approaches for nearest neighbor search \cite{DBLP:journals/is/AumullerBF20}, where different feature subsets or combinations could be relevant for different patterns or locations in a dataset.

There are different aspects of feature selection algorithms that can be investigated, such as performance, amount of dimensionality reduction, stability, and complexity \cite{DBLP:journals/cee/ChandrashekarS14}. A combination of these aspects can also be considered as the target of the conducted evaluations. In most cases, feature selection algorithms are evaluated using supervised measures assessing the quality of downstream tasks such as classification or clustering. There are however some metrics proposed specifically for the evaluation of feature selection algorithms \cite{DBLP:journals/algorithms/MostertME21}, yet these metrics also only investigate limited aspects of feature selection algorithms. On the other hand, feature selection stability metrics \cite{DBLP:conf/pkdd/Nogueira016, DBLP:conf/aia/Kuncheva07} attempt to quantify whether the same features are selected in different scenarios (e.g., presence of noise) or not. Although these metrics can present valuable insights, their assumptions, in some cases, can render them ineffective for stability analysis. For instance, there might be two different subsets of features selected by an algorithm which represent the same information (e.g., distance in kilometers and distance in miles). In this case, the algorithm will be categorized as unstable while it is perfectly stable based on the information included in the features.

In this paper, we propose Feature Selection Dynamic Evaluation Metric (FSDEM)\footnote{The implementation of FSDEM and its corresponding stability score is available at: https://github.com/mrajabinasab/FSDEM-Feature-Selection-Dynamic-Evaluation-Metric}. FSDEM is focused on assessing the performance and stability of feature selection methods. FSDEM offers a dynamic framework for the assessment and evaluation of the feature selection algorithm which can be instantiated with any performance measure. Hence, it is able to give insights on different performance aspects of the algorithm. It also yields insights into the stability of the feature selection algorithm. Unlike previous methods, FSDEM defines stability based on the changes in the performance measure as the number of selected features varies. Therefore, it avoids the challenges posed by the previous methods discussed above.

This paper is an extended version of a paper published at 17th International Conference on Similarity Search and Applications (SISAP 2024) \cite{rajabinasab2024}. The remainder of the paper is structured as follows: In Section~\ref{sec:related_works} we review previous works in metrics and assessment criteria proposed for the evaluation of feature selection algorithms. In Section~\ref{sec:proposed_method}, we discuss the proposed method ``FSDEM'' in detail. In Section~\ref{sec:exp}, we present analysis and empirical results to demonstrate the performance of the proposed feature selection evaluation metric. Finally, in Section~\ref{sec:conc}, we conclude the paper and discuss possible future directions.

\section{Related Work}  \label{sec:related_works}
To the best of our knowledge, there are only few papers proposing metrics for the evaluation of different characteristics of feature selection algorithms. We can divide feature selection algorithms based on using or not using labels to carry out feature selection tasks into supervised and unsupervised methods. The evaluation of feature selection algorithms based on the evaluation of downstream tasks however is typically supervised (or based on so-called external quality measures). Additionally, some metrics focus on special aspects of the feature selection problem more directly.
%In this section, we also review two papers among which that propose a novel approach for evaluating different aspects of the feature selection algorithms.

\subsection{Evaluation Based on a Downstream Task}
Feature selection algorithms are usually evaluated based on their performance in a downstream machine learning or data mining task. The downstream task can be supervised (e.g., classification) or unsupervised (e.g., clustering). 

Evaluation of feature selection algorithms w.r.t.\ a supervised downstream task \cite{DBLP:journals/nca/GuhaKSSB21, DBLP:journals/pr/FanLTLLD24, DBLP:journals/asc/PanCX23, DBLP:journals/asc/LiuLSL24, DBLP:journals/istr/AhmadYLCLUSWHLZSC24} involves using one or more classifiers to investigate how different classification performance measures change after employing feature selection. Supervised feature selection algorithms use this approach and measures such as accuracy, precision, recall, F1-score, and AUC to show the efficacy of the feature selection procedure for supervised tasks. Despite valuable insights extracted from these evaluations and their respective experimental results, they are highly dependant on the selection of the classifier and its characteristics. 

While unsupervised feature selection algorithms are sometimes also evaluated by supervised downstream tasks \cite{DBLP:journals/prl/ZhengZWZYG20, DBLP:journals/tip/ShiZLZC23}, the evaluation based on an unsupervised downstream task is more common for unsupervised feature selection methods. This is usually performed using performance measures associated with unsupervised machine learning tasks \cite{10123301, DBLP:journals/ijon/ShangKWZWLJ23, DBLP:journals/prl/WANG2024110183, LI2024120227, DBLP:journals/ijon/JahaniAES23, DBLP:journals/kbs/CaoXSQ23}. However, while the downstream task is unsupervised, the employed performance measures, such as clustering accuracy, Normalized Mutual Information (NMI), and purity, are themselves supervised (or external) measures.  Despite using an unsupervised downstream task, these measures are calculated based on the agreement between clustering results and ground truth labels existing in the datasets. Therefore these evaluations are also strongly connected to the characteristics of the underlying unsupervised algorithm which is employed to conduct them.

\subsection{Specialized Evaluation Measures}
Some metrics have been proposed that evaluate specific aspects of a feature selection algorithm. These metrics address the stability of the feature selection algorithm by investigating its behavior in different scenarios \cite{DBLP:conf/pkdd/Nogueira016, DBLP:conf/aia/Kuncheva07} and the effectiveness of a feature selection algorithm in increasing the accuracy while reducing the number of features \cite{DBLP:journals/algorithms/MostertME21}. 

\subsubsection{Stability Measures}
Nogueira et~al.~\cite{DBLP:conf/pkdd/Nogueira016} propose a metric to calculate the stability of a feature selection algorithm in the presence of noise. This metric is based on Pearson's correlation coefficient.
In this metric, the optimal value is 1, the minimal value is -1, and the constant value for a random feature selection algorithm is 0. The metric is given by:
\begin{equation}\label{eq:SM}
\hat{\Phi} (Z) = 1 - \frac{\frac{1}{d} \sum_{f=1}^{d} s_f^2} {\frac{\Bar{k}}{d}(1 - \frac{\Bar{k}}{d})}
\end{equation}
where $d$ is the number of features in the dataset, $\Bar{k}$ is the number of selected features, and $s_f^2$ is the unbiased sample variance of the selection of the $f^{\mathit{th}}$ feature. $\hat{\Phi} < 0.40$ is considered to indicate poor stability, whereas $0.40 \leq \hat{\Phi} \leq 0.75$ imply an intermediate to good level of stability, and $\hat{\Phi} > 0.75$ suggests the algorithm to be nearly perfectly stable. 

An alternative stability measure \cite{DBLP:conf/aia/Kuncheva07} puts the focus on Sequential Forward Selection (SFS) considering the intersection between two sets of selected features to measure the stability. Let $S_1$, $S_2$,...,$S_K$ be the sequences of features obtained from $K$ runs of SFS on a given dataset. The stability index proposed by Kuncheva \cite{DBLP:conf/aia/Kuncheva07} for a set of sequence of features $A = \{S_1, S_2,...,S_K\}$ for a given set of size $k$ is given by:
\begin{equation}\label{eq:SM2}
\mathcal{I}_S(A(k)) = \frac{2}{K(K-1)} \sum_{i=1}^{K-1} \sum_{j=i+1}^{K} I_C(S_i(k), S_j(k))
\end{equation}
where $I_C$ is the consistency index \cite{DBLP:conf/aia/Kuncheva07}, a measure of the similarity between two selected feature subsets.

\subsubsection{Baseline Fitness Improvement}
The Baseline Fitness Improvement (BFI) measure \cite{DBLP:journals/algorithms/MostertME21} is a normalized metric for the evaluation and comparison of feature selection algorithms which quantifies the potential value gained by applying feature selection. BFI is given by:
\begin{equation}\label{eq:BF}
\BFI(s) = f(s) - f(s^*)
\end{equation}
where $f(s)$ is the fitness of the selected feature subset and $f(s^*)$ is the baseline fitness (fitness of complete feature set), defined as: 
\begin{equation}\label{eq:FF}
f(s) = k_c \cdot f_c(s) + k_p \cdot f_p(s)
\end{equation}
where $f_c(s)$ is some classifier performance function and $f_p(s)$ is a penalty function which grows exponentially as the number of selected features increases.

\subsection{Discussion}
It is evident that the current evaluation metrics for feature selection algorithms do not completely address the assessment of what we expect of a feature selection algorithm (ideally, the improvement of the accuracy and the reduction of dimensionality). Evaluation based on a downstream task has some limitations. For instance, (classification) accuracy only reflects the classification performance after the selection procedure without considering the number of selected features and the effect of adding or removing one feature. On the other hand, clustering accuracy only illustrates the performance of the selected features in carrying out the clustering task. Hence, it suffers from the same shortcomings. 

The stability measure \cite{DBLP:conf/pkdd/Nogueira016} investigates the effect of noise on the feature selection output and does not provide analysis of the confidence of the feature selection algorithm. BFI is the more consistent approach for evaluating feature selection algorithms, but still, it does not address the advantage gain by adding or removing a feature. It also does not provide insights into how stable the performance of the feature selection algorithm is with different numbers of features to be selected. On the other hand, there are usually confounding effects involved with real data. The stability measure \cite{DBLP:conf/pkdd/Nogueira016} is not able to capture their impact on the feature selection procedure as they are considered to be different features while they might include the same amount of information. All of these challenges and shortcomings motivate this study to propose a new feature selection algorithm which can better reflect the blind spots of a feature selection method.

\section{Proposed Method} \label{sec:proposed_method}
To address some of the challenges and shortcomings described above, here we propose Feature Selection Dynamic Evaluation Metric (FSDEM). FSDEM is dynamic and it can be integrated with any performance measure to provide insights to the feature selection algorithm.  FSDEM has two properties that can be effective for the analysis of different aspects of a feature selection method. In the remainder of this section, we discuss the two properties of FSDEM.

\subsection{FSDEM Score}
Let $F$ be the number of features. Let $M(f)$ be the value of an arbitrary evaluation performance measure with $f$ out of $F$ features selected. Let $g(x)$ be the function approximated from $M(f)$ with different observations of $f \in \{1,...,F\}$. The FSDEM score will be the area under the curve of $g(x)$:
\begin{equation}\label{eq:FSDEM}
\FSDEM = \frac{\int_{a}^{b} g(x) \,dx}{b-a}
\end{equation}

Clearly, the range of the possible values for FSDEM score is bounded, and it inherits its bounds from the measure $M$. Assuming that the values of the measure $M$ are in the range $[0, 1]$ (bounds of most ordinary measures such as accuracy), we desire to have values closer to 1 as it indicates a better performance achieved by the feature selection algorithm. 

There are three main advantages to using the FSDEM score: \textit{i}) If the number of features is large, we can run fewer experiments and have the curve with a certain degree of precision. \textit{ii}) Some feature selection algorithms might perform better in selecting a specific number of features for the target machine learning or data mining task. Using FSDEM, we can set $a$ and $b$ according to our preferred range and hence, select a feature selection method which performs best regarding our preferences. \textit{iii}) We can use the first-order derivative as a measure of stability. 

\subsection{Stability Score}
The stability score is calculated based on the first-order derivative of~\textit{g}:
\begin{equation}\label{eq:S1}
  \STAB = \frac{\sum_{i=a}^{b} g'(x)}{(b-a) + 1}
\end{equation}
The first-order derivative of $g$ indicates the amount of change in the performance measure w.r.t.\ different number of features. As a result, the stability score can be interpreted as a value in the range $[-1, 1]$ (for the measures with values in range $[0, 1]$). Values close to $-1$ indicate that the change in the evaluation measure tends to be negative, and adding more features reduces the performance. Values close to $1$ indicate that the change in the evaluation measure is predominantly positive, and it is beneficial, w.r.t.\ the performance measure used to approximate the function, to include more features selected by the feature selection method. A value close to $0$ indicates that either the performance measure fluctuates a lot as the number of selected features changes or that adding more features has a negligible impact on the evaluation metric. In both cases, it indicates that the feature selection algorithm performs poorly based on the selected performance measure used to approximate the function. Ideally, we desire to have values closer to 1 for the stability score since it indicates a better selection of features. It is noteworthy that in real scenarios, the values are often very close to 0 as the performance measures do not change dramatically in the value with different observations. Hence, positive values are considered to be good in general. Also the stability value can be used effectively in the comparisons alongside with FSDEM to select the best feature selection algorithm for a given task and dataset.

The stability metric proposed in this paper accounts for the informative value of the selected features and not whether the same features are selected or not. Hence, it does not suffer from the assumption of the previous methods. For example, consider a scenario where a feature selection algorithm selects two feature subsets $\digamma = \{f_1, f_2, ..., f_m\}$ and $\digamma' = \{f'_1, f'_2, ..., f'_m\}$ on two different noisy versions of dataset $D$. If  $\digamma \cap \digamma' = \emptyset$, the algorithm is considered to be totally unstable based on the stability defined by the previous metrics (Equations~\ref{eq:SM} and \ref{eq:SM2}). Whereas $\digamma_1$ and $\digamma_2$ might represent the same information or include the same amount of informative value. FSDEM stability score is calculated based on the effect of selected feature subset on the final downstream task and hence, it can successfully identify stable or unstable feature selection algorithms based on the informative value of the selected features.

\subsection{Implementation and Instantiation} 
In order to approximate a continuous function from a set of discrete observations we use linear approximation. Linear approximation is the simplest yet most efficient method for approximating a continuous function from the observations of different discrete observations of the performance measure. Linear approximation simply draws a line between two consequent discrete observations. The trapezoidal rule \cite{atkinson1991introduction} is used to calculate the integration. The finite difference method \cite{Co_2013} is employed in order to calculate the first-order derivative. 

\begin{figure}[tb!]
 \center
  \includegraphics[width=\textwidth]{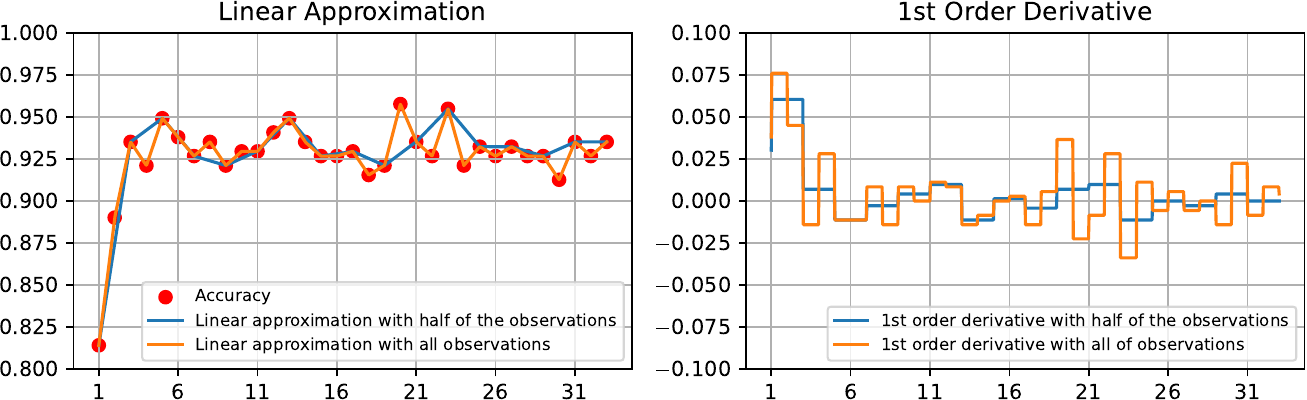}
  \caption{Example of the function approximation (left) and first order derivative (right) with all and half of the observation, based on the accuracy as the performance measure.}
  \label{fg:fg1}
\end{figure}

FSDEM can be integrated with any arbitrary evaluation metric. Fig.~\ref{fg:fg1} illustrates an example of the function approximation and its first-order derivative as the backbone of the FSDEM and the stability score based on the accuracy as the selected performance measure. The two cases shown are for the approximation with half and all of the observations. As shown in Fig.~\ref{fg:fg1}, even when approximating the function with half of the observations, both results show adequate consistency.

% \subsection{2nd Order Stability Score:}
% The second-order stability score is calculated based on the second-order derivative of \textit{g}. The formula for calculation of this stability score is mentioned in eq.~(\ref{eq:S2}).
    
% \begin{equation}\label{eq:S2}
%      S_2 = \frac{\sum_{i=1}^{F} g''(x)}{F} 
% \end{equation} 

% The second-order derivative of \textit{g} indicates the amount of change in the 1st order stability score (the amount of change in the amount of change of the evaluation metric) observed by increasing the number of selected features. As a result, 2nd order stability score can be interpreted as a value in range $(-1, 1)$.

\section{Analysis and Empirical Results}  \label{sec:exp}
In this section, we present the analysis and empirical results to provide insights into the proposed feature selection evaluation metric. First, we present the capabilities of FSDEM in identifying the best feature selection algorithm w.r.t.\ the desired range for the number of selected features. Then, we investigate the stability score of FSDEM by analyzing a case in which the more stable feature selection algorithm is identified by FSDEM and not by the previous stability metric \cite{DBLP:conf/pkdd/Nogueira016}. 

We provide empirical results for the FSDEM and its stability score. We use 20 different datasets to conduct the experiments. The details of the datasets and their characteristics are presented in Table~\ref{tbl:ds}. All of the datasets are available on the UCI repository \cite{uci_ml_repository}. Random feature selection is used as the baseline method for the experiments. Information gain, chi2, and a wrapper-based feature selection based on random forest are the three other feature selection algorithms we use for the experiments.

\begin{table}[tb!]
\caption{Selected datasets for the experiments and their characteristics.}\label{tbl:ds}
\centering
\begin{tabular}{rrrr}
\toprule
Dataset Key &  Name & \# of Instances & \# of Features \\
\midrule
D1 & audiology & 200 & 70 \\
D2 & automobile & 205 & 25 \\
D3 & breast-cancer & 569 & 30 \\
D4 & credit & 690 & 15 \\
D5 & cylinder & 541 & 39 \\
D6 & dermatology & 366 & 34 \\
D7 & glass & 214 & 9 \\
D8 & hepatitis & 155 & 19 \\
D9 & horse-colic & 368 & 27 \\
D10 & image & 210 & 19 \\
D11 & ionosphere & 351 & 34 \\
D12 & iris & 150 & 4 \\
D13 & lung-cancer & 32 & 56 \\
D14 & lymphography & 148 & 19 \\
D15 & primary-tumor & 339 & 17 \\
D16 & raisin & 900 & 7 \\
D17 & tic-tac-toe & 958 & 9 \\
D18 & wine & 178 & 13 \\
D19 & yeast & 1484 & 8 \\
D20 & zoo & 101 & 16 \\
\bottomrule
\end{tabular}
\end{table}

\subsection{Analysis}
In this section, we present two cases to analyze different properties of FSDEM. First, we use dataset D6 and two feature selection methods, namely information gain and chi2 to show that the best feature selection algorithm can vary based on the target size of the selected feature subset and that FSDEM can successfully identify this. Second, we present a case in which the assumption of previous feature selection stability measures renders them ineffective, while FSDEM stability works well.

\subsubsection{Evaluation based on the target number of features:} 
\begin{figure}[tb!]
 \center
  \includegraphics[width=\textwidth]{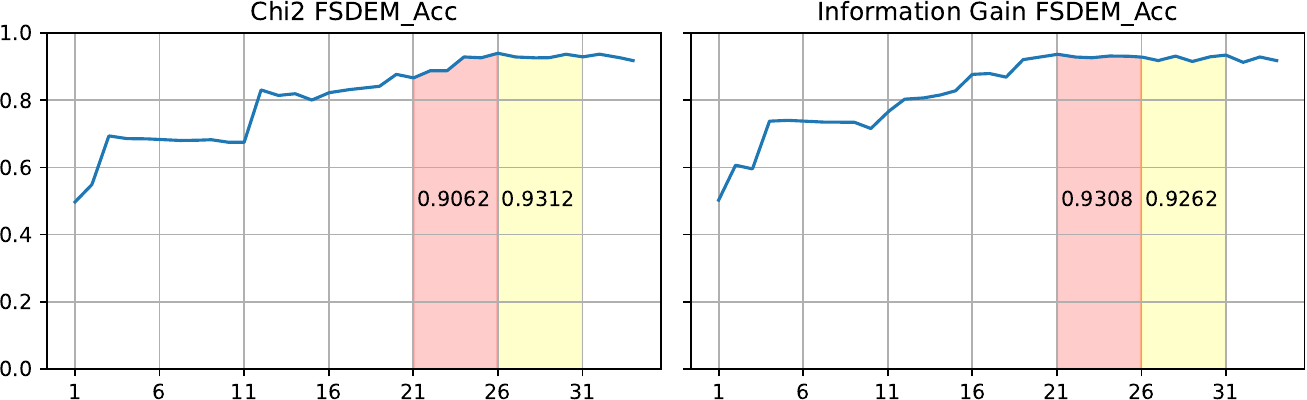}
  \caption{FSDEM score based on accuracy for two different feature selection methods and target feature number ranges.}
  \label{fg:figta1}
\end{figure}

Fig.~\ref{fg:figta1} shows the accuracy curve for chi2 and information gain on D6. The FSDEM score for different ranges of target feature numbers is presented in the figure. For a target feature number in the range $[21, 26]$, information gain is a better choice as it yields a higher overall accuracy. But for a target feature number in the range $[26, 31]$, chi2 performs better.

FSDEM successfully provides insights based on the target number of features and can contrast different feature selection methods in this respect. This enables selecting the best method based on a specific number of selected features.

\subsubsection{Stability score based on the informative value:} As discussed earlier, previous methods proposed for investigating the stability of feature selection algorithms measure stability as the ability to select the same set of features in different scenarios \cite{DBLP:conf/pkdd/Nogueira016}. However, in some cases, the same information might be included in different features. A feature selection algorithm that selects different features in different scenarios is considered unstable by these methods, regardless of the information included in the features.

To investigate this challenge a bit further, we create a dummy dataset. This dataset consists of 500 samples and 6 features. The features include: \textit{\{age, salary in EUR, salary in USD, size of the residence, distance to the city center in kilometers, distance to the city center in miles\}}. The target variable is \textit{wealth}, which is a binary variable that indicates whether a person is wealthy or not. 
Consider a feature selection algorithm $A$. The feature rankings calculated by $A$ and two different scenarios are \textit{\{salary in EUR, size of the residence, distance to the city center in kilometers, age, salary in USD, distance to the city center in miles\}} and \textit{\{salary in USD, size of the residence, distance to the city center in miles, age, salary in EUR, distance to the city center in kilometers\}}.

\begin{figure}[tb!]
 \center
  \includegraphics[width=\textwidth]{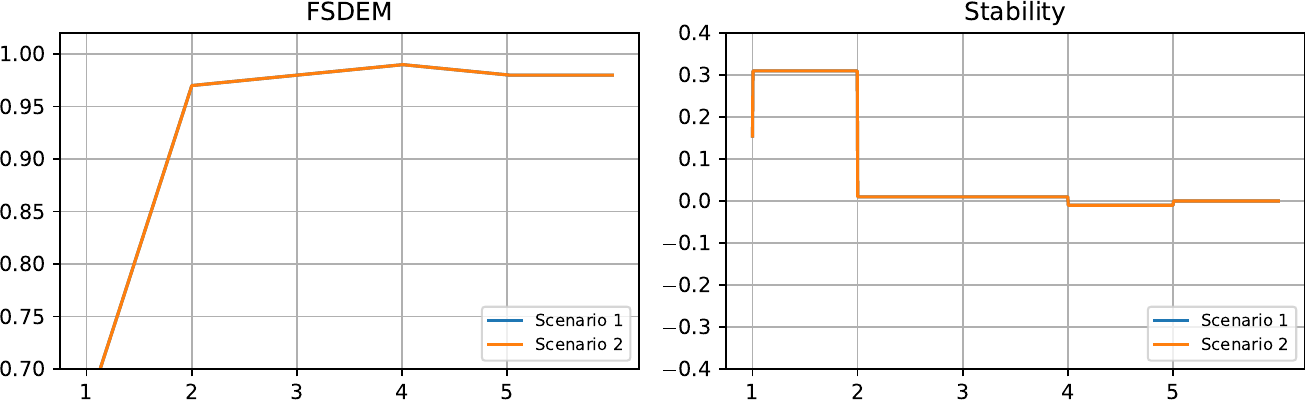}
  \caption{FSDEM and stability score for the two scenarios.}
  \label{fg:figta2}
\end{figure}

If we aim to select half of the features (3 in this case) for a machine learning or data mining task, the stability measure proposed by Noguiera et al.~\cite{DBLP:conf/pkdd/Nogueira016} yields a value of -0.3333 (Eq.~\ref{eq:SM}). This value is considered indicative of poor stability. However, the information represented in both of the feature subsets is the same. So from an information-theoretic point of view, the algorithm must be considered stable as it is able to select the same amount of information in both cases. In~Fig.~\ref{fg:figta2}, an illustration of the FSDEM and its stability score for the two scenarios is shown. 

As FSDEM takes performance into account, we include figures for both scenarios. Evidently, the algorithm shows the same values in both scenarios. The average value for the stability score of the two scenarios is 0.0466 which is a positive value and indicates good stability w.r.t.\ accuracy. 

FSDEM relies on the integrated performance measure as a factor for stability. In this case, despite different features being selected, the feature subsets represent the same information and hence the underlying feature selection algorithm must not be considered unstable. FSDEM successfully captures the informative aspect of stability instead of strictly requiring the same features to be selected, which allows for more robust measures in data with confounding effects.

\subsection{Empirical Results}
\begin{figure}[tb!]
 \center
  \includegraphics[width=\textwidth]{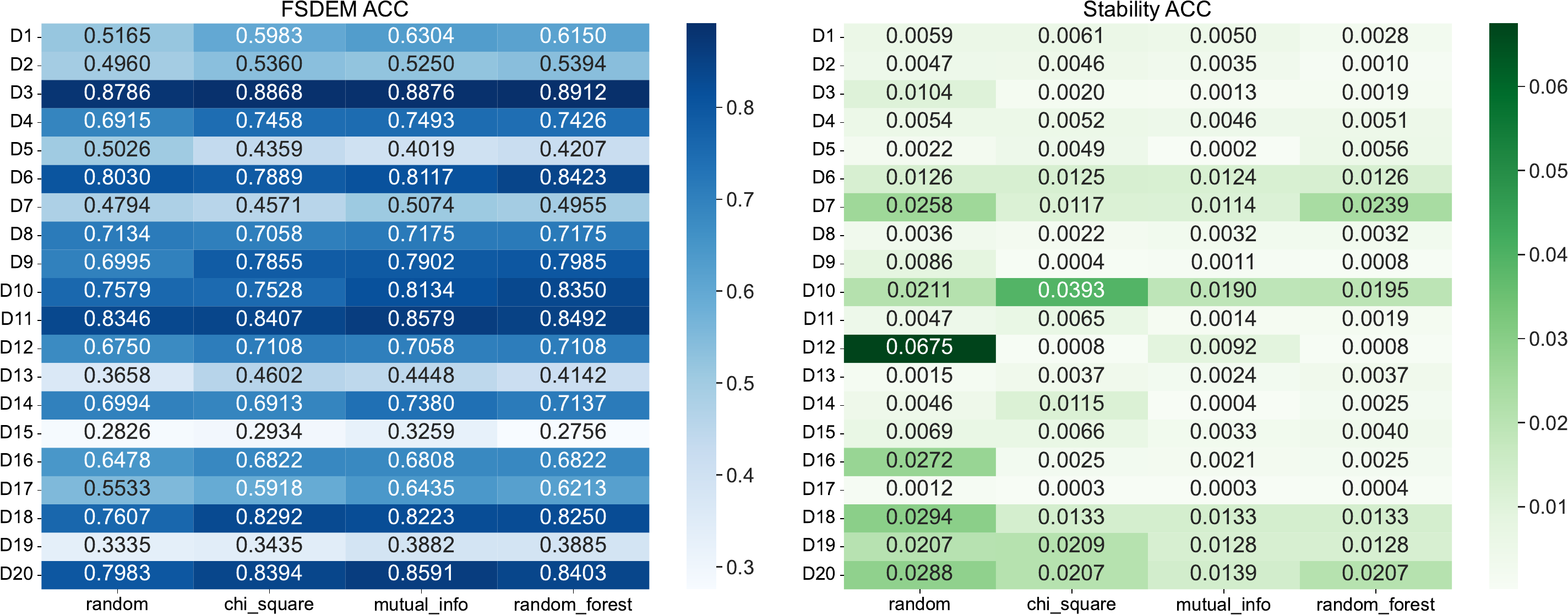}
  \includegraphics[width=\textwidth]{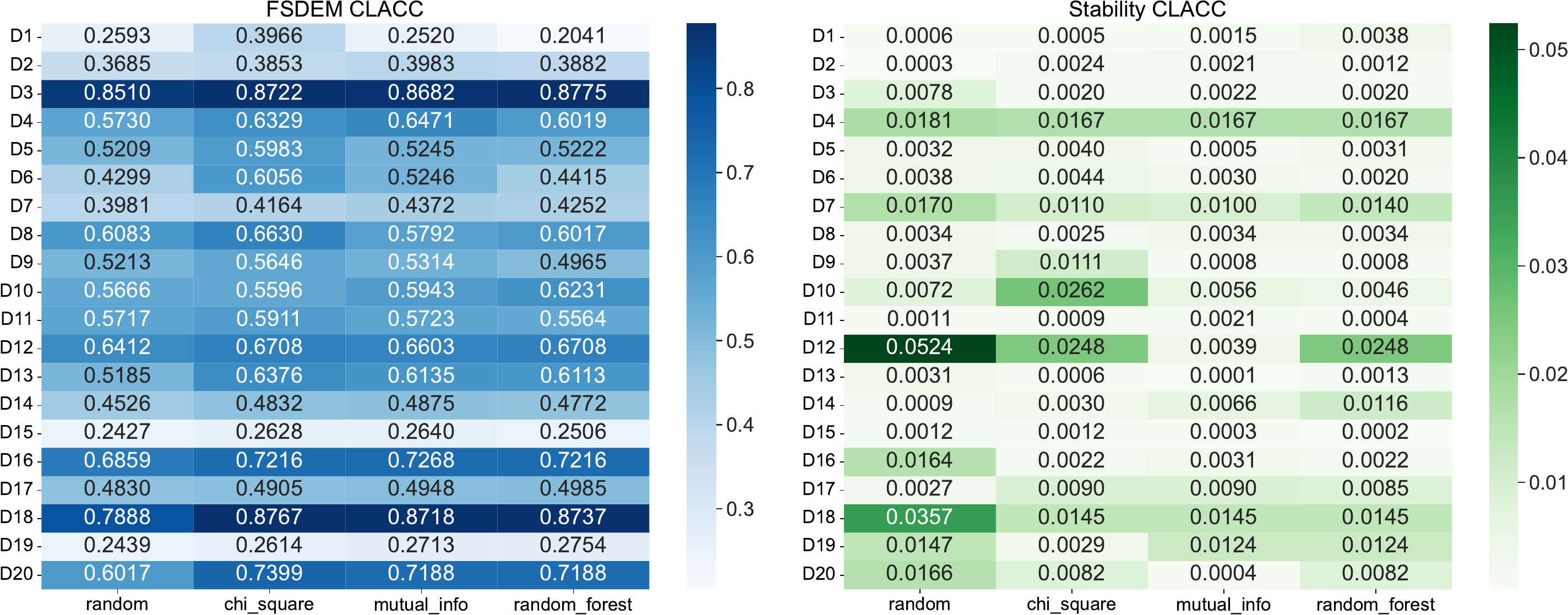}
  \caption{FSDEM score and its associated stability for different algorithms and datasets based on accuracy and clustering accuracy. Note that the ranges are different.}
  \label{fg:fg2}
\end{figure}

In this section, we provide empirical results to illustrate the effectiveness of FSDEM and its stability score in the successful evaluation of feature selection algorithms. In Fig.~\ref{fg:fg2}, we display the experimental results in terms of FSDEM score and its associated stability value for two different performance measureS, namely (classification) accuracy and clustering accuracy. 

We use two different performance measureS to show that FSDEM can successfully integrate different measures to provide insights into the feature selection algorithms' performance. Clustering Accuracy (CLACC) \cite{DBLP:conf/cvpr/CaoZFLZ15}, defined as:
\begin{equation}\label{eq:CLACC}
\CLACC = \frac{1}{n} \sum_{i=1}^{n} \delta(c_i, \text{map}(g_i))
\end{equation}
calculates the accuracy of clustering based on the clustering label $c_i$ and the true label (i.e., the class label) $g_i$, $\text{map}(.)$ is an optimal mapping function which utilizes the Hungarian method \cite{DBLP:books/ph/PapadimitriouS82} to match the clustering labels with true labels. The indicator function, $\delta$,  returns 1 if its inputs are equal and 0 otherwise. 

The experimental results shown in Fig.~\ref{fg:fg2} indicate that, as expected, the wrapper-based feature selection algorithm based on random forest has the best overall performance compared to the other algorithms. The stability results show that the most stable algorithm in feature selection is the random feature selection procedure. This is expected as in random feature selection, a feature is selected without considering any specific criteria. Features are selected randomly at every step. Hence, by selecting more features, it is most likely that we can include the informative features that are important for the final task. As a result, the final performance is improved by increasing the number of selected features. On the other hand, when it comes to the other methods, it might be more beneficial to select fewer features as it is more likely that the most informative features are selected in the first steps. Investigating the values of the FSDEM score and its associated stability allows one to select the most efficient and stable feature selection algorithm for an arbitrary task.

\begin{figure}[tb!]
 \center
  \includegraphics[width=\textwidth]{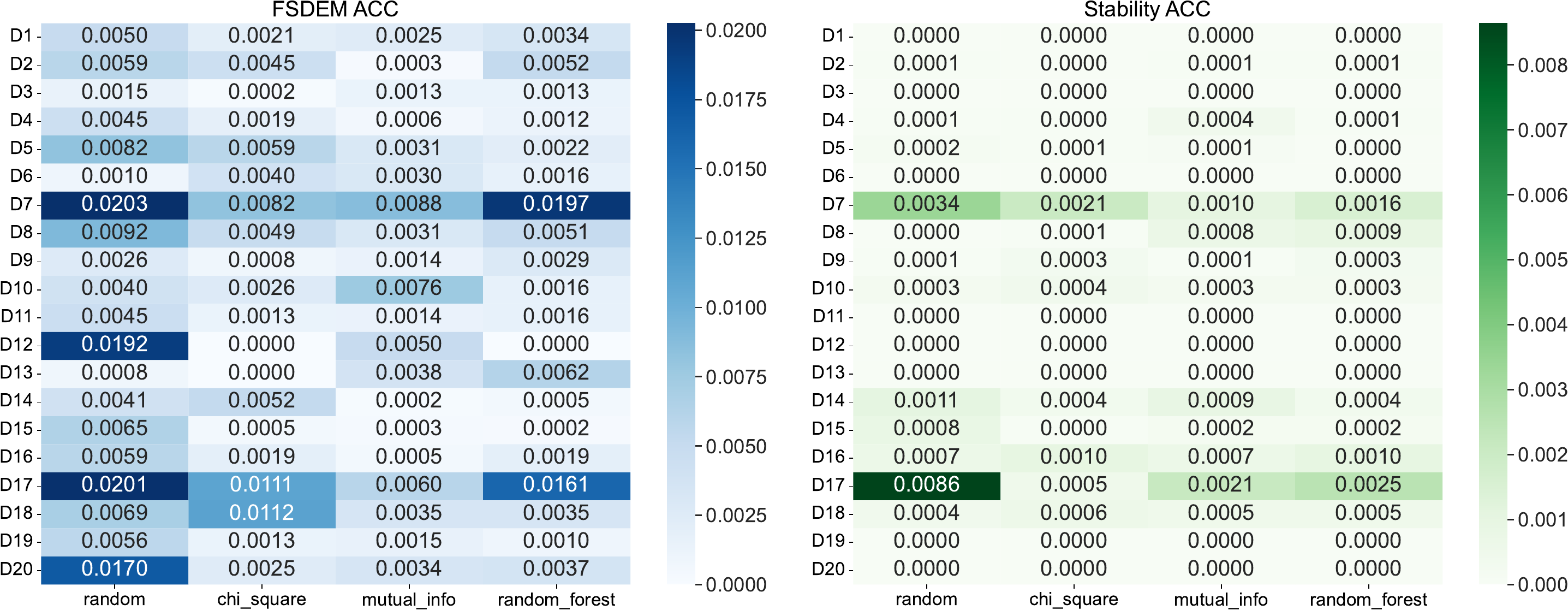}
  \includegraphics[width=\textwidth]{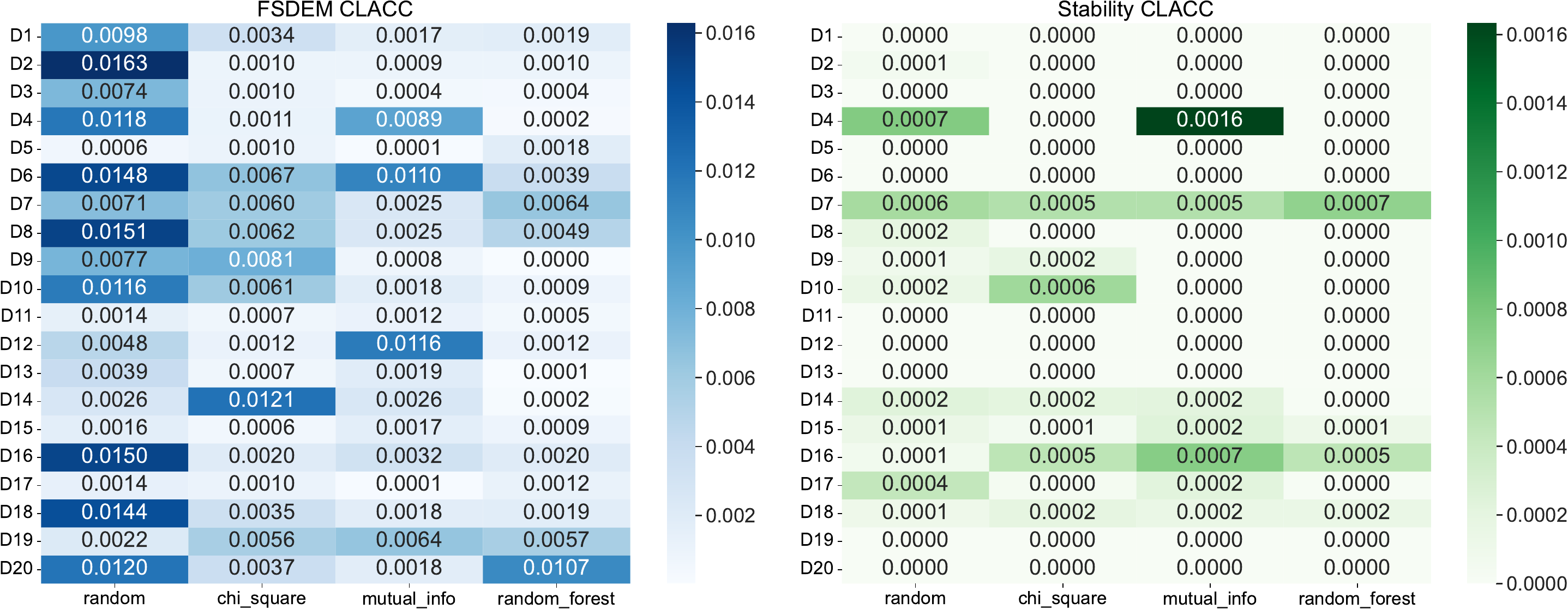}
  \caption{Absolute difference of FSDEM score and its associated stability with all and half of the observations. Note that the ranges are different.}
  \label{fg:fg3}
\end{figure}

Fig.~\ref{fg:fg3} illustrates the absolute difference between the FSDEM score and its associated stability with all and half of the observations for each case. It is clear that the margin of error in both cases is low. Hence, it is feasible to use FSDEM and its associated stability score with a high degree of confidence and fewer actual observations required. This makes FSDEM effective for conducting a thorough investigation on high-dimensional datasets and slow feature selection algorithms.

\section{Conclusion}  \label{sec:conc}
In this paper, we proposed FSDEM, a novel dynamic metric to evaluate the performance and stability of feature selection algorithms. FSDEM has the capability to integrate any performance measure to provide insights into different aspects of the performance of a feature selection algorithm and the stability associated with it. 

FSDEM is able to provide a complete and thorough analysis of a feature selection algorithm w.r.t.\ its performance due to different numbers of selected features. Additionally, FSDEM can help to select the best feature selection algorithm for a desirable target range for the number of features. The stability score provided by FSDEM can assess the stability of a feature selection algorithm w.r.t.\ the informative value of the selected feature subset. It can also reflect the overall effectiveness of a feature selection algorithm. The conducted analysis and presented empirical results of the evaluations illustrate that FSDEM is an effective measure for evaluation of the performance and stability of a feature selection algorithm. Empirical results also suggest that FSDEM can work efficiently with a limited number of observations and, hence, is effective for the experiments involved with high-dimensional datasets or slow feature selection algorithms.

Despite of all the strengths and capabilities of FSDEM, there is a limitation related to its stability score. As the stability score depends on a function approximated from a performance measure which in most cases ranges between 0 and 1, it often outputs values close to 0. As the most important use-case for the evaluation metrics is the comparison, it might not be considered as a severe problem. However, methods such as correction for chance can be applied in order to adjust the value of the stability score. This can be the main focus of future studies conducted in this area.
\printbibliography
\end{document}